# Efficient Reinforcement Learning Using
# Recursive Least-Squares Methods

**Xin Xu**                                                    XUXIN_MAIL@263.NET
**Han-gen He**                                               HEHANGEN@CS.HN.CN
**Dewen Hu**                                                 DWHU@NUDT.EDU.CN
*Department of Automatic Control*
*National University of Defense Technology*
*ChangSha, Hunan, 410073, P.R.China*

## Abstract

The recursive least-squares (RLS) algorithm is one of the most well-known algorithms used in adaptive filtering, system identification and adaptive control. Its popularity is mainly due to its fast convergence speed, which is considered to be optimal in practice. In this paper, RLS methods are used to solve reinforcement learning problems, where two new reinforcement learning algorithms using linear value function approximators are proposed and analyzed. The two algorithms are called RLS-TD($\lambda$) and Fast-AHC (Fast Adaptive Heuristic Critic), respectively. RLS-TD($\lambda$) can be viewed as the extension of RLS-TD(0) from $\lambda = 0$ to general $0 \le \lambda \le 1$, so it is a multi-step temporal-difference (TD) learning algorithm using RLS methods. The convergence with probability one and the limit of convergence of RLS-TD($\lambda$) are proved for ergodic Markov chains. Compared to the existing LS-TD($\lambda$) algorithm, RLS-TD($\lambda$) has advantages in computation and is more suitable for online learning. The effectiveness of RLS-TD($\lambda$) is analyzed and verified by learning prediction experiments of Markov chains with a wide range of parameter settings.

The Fast-AHC algorithm is derived by applying the proposed RLS-TD($\lambda$) algorithm in the critic network of the adaptive heuristic critic method. Unlike conventional AHC algorithm, Fast-AHC makes use of RLS methods to improve the learning-prediction efficiency in the critic. Learning control experiments of the cart-pole balancing and the acrobot swing-up problems are conducted to compare the data efficiency of Fast-AHC with conventional AHC. From the experimental results, it is shown that the data efficiency of learning control can also be improved by using RLS methods in the learning-prediction process of the critic. The performance of Fast-AHC is also compared with that of the AHC method using LS-TD($\lambda$). Furthermore, it is demonstrated in the experiments that different initial values of the variance matrix in RLS-TD($\lambda$) are required to get better performance not only in learning prediction but also in learning control. The experimental results are analyzed based on the existing theoretical work on the transient phase of forgetting factor RLS methods.

## 1. Introduction

In recent years, reinforcement learning (RL) has been an active research area not only in machine learning but also in control engineering, operations research and robotics (Kaelbling et al.,1996; Bertsekas, et al.,1996; Sutton and Barto,1998; Lin,1992). It is a computational approach to understand and automate goal-directed learning and decision-making, without relying on exemplary supervision or complete models of the environment. In RL, an agent is placed in an initial unknown environment and only receives evaluative feedback from the environment. The feedback is called reward or reinforcement signal. The ultimate goal of RL is to learn a strategy for selecting actions such that the expected sum of discounted rewards is maximized.

Since lots of problems in the real world are sequential decision processes with delayed evaluative feedback, the research in RL has been focused on theory and algorithms of learning to solve the optimal control problem of Markov decision processes (MDPs) which provide an elegant mathematical model for sequential decision-making. In operations research, many results have been presented to solve the optimal control problem of MDPs with model information. However, in reinforcement learning, the model information is assumed to be unknown, which is different from the methods studied in operations research such as dynamic programming. In dynamic programming, there are two elemental processes, which are the policy evaluation process and the policy improvement process, respectively. In RL, there are two similar processes. One is called learning prediction and the other is called learning control. The goal of learning control is to estimate the optimal policy or optimal value function of an MDP without knowing its model. Learning prediction aims to solve the policy evaluation problem of a stationary-policy MDP without any prior model and it can be regarded as a sub-problem of learning control. Furthermore, in RL, learning prediction is different from that in supervised learning. As pointed out by Sutton (1988), the prediction problems in supervised learning are single-step prediction problems while those in reinforcement learning are multi-step prediction problems. To solve multi-step prediction problems, a learning system must predict outcomes that depend on a future sequence of decisions. Therefore, the theory and algorithms for multi-step learning prediction become an important topic in RL and much research work has been done in the literature (Sutton, 1988; Tsitsiklis and Roy, 1997).

Among the proposed multi-step learning prediction methods, temporal-difference (TD) learning (Sutton, 1988) is one of the most popular methods. It was studied and applied in the early research of machine learning, including the celebrated checkers-playing program (Minsky, 1954; Samuel, 1959). In 1988, Sutton presented the first formal description of temporal- difference methods and the TD($\lambda$) algorithm (Sutton,1988). Convergence results are established for tabular temporal-difference learning algorithms where the cardinality of tunable parameters is the same as that of the state space (Sutton, 1988; Watkins,et al.,1992; Dayan,et al., 1994; Jaakkola, et al.,1994). Since many real-world applications have large or infinite state space, value function approximation (VFA) methods need to be used in those cases. When combined with nonlinear value function approximators, TD($\lambda$) can not guarantee convergence and several results regarding divergence have been reported in the literature (Tsitsiklis and Roy,1997). For TD($\lambda$) with linear function approximators, also called linear TD($\lambda$) algorithms, several convergence proofs have been presented. Dayan (1992) showed the convergence in the mean for linear TD($\lambda$) algorithms with arbitrary $0 \leq \lambda \leq 1$. Tsitsiklis and Roy (1994) proved the convergence for a special class of TD learning algorithms, known as TD(0), while in Tsitsiklis and Roy (1997), they extended the early results to general linear TD($\lambda$) case and proved the convergence with probability one.

The above linear TD($\lambda$) algorithms have rules for updating parameters similar to those in gradient-descent methods. However, as in gradient-learning methods, a step-size schedule must be carefully designed not only to guarantee convergence but also to obtain good performance. In addition, there is inefficient use of data that slows the convergence of the algorithms. Based on the theory of linear least-squares estimation, Brartke and Barto (1996) proposed two temporal-difference algorithms called the Least-Squares TD(0) algorithm (LS-TD(0)) and the Recursive Least- Squares TD(0) algorithm (RLS-TD(0)), respectively. LS-TD(0) and RLS-TD(0) are more efficient in a statistical sense than conventional linear TD($\lambda$) algorithms and they eliminate the usage of step-size schedules. Furthermore, the convergence of LS-TD(0) and RLS-TD(0) has been provided in theory. The above two algorithms can be viewed as the least-squares versions of conventional linear TD(0) methods. However, as has been shown in the literature, TD learning algorithms such as TD($\lambda$) with $0<\lambda<1$ that update predictions based on the estimates of multiple steps are more efficient than Monte-Carlo methods as well as TD(0). By employing the mechanism of eligibility traces, which is determined by $\lambda$, TD($\lambda$) algorithms with $0<\lambda<1$ can extract more information from historical data. Recently, a class of linear temporal-difference learning algorithms called LS-TD($\lambda$) has been proposed by Boyan (1999,2002), where least-squares methods are employed to compute the value-function estimation of TD($\lambda$) with $0\leq\lambda\leq1$. Although LS-TD($\lambda$) is more efficient than TD($\lambda$), it requires too much computation per time-step when online updates are needed and the number of state features becomes large.

In system identification, adaptive filtering and adaptive control, the recursive least-squares (RLS) (Young,1984; Ljung, 1983; Ljung,1977) method, commonly used to reduce the computational burden of least-squares methods, is more suitable for online estimation and control. Although RLS-TD(0) makes use of RLS methods, it does not employ the mechanism of eligibility traces. Based on the work of Tsitsiklis and Roy (1994, 1997), Boyan (1999,2002) and motivated by the above ideas, a new class of temporal-difference learning methods, called the RLS-TD($\lambda$) algorithm, is proposed and analyzed formally in this paper. RLS-TD($\lambda$) is superior to conventional linear TD($\lambda$) algorithms in that it makes use of RLS methods to improve the learning efficiency in a statistical point of view and eliminates the step-size schedules. RLS-TD($\lambda$) has the mechanism of eligibility traces and can be viewed as the extension of RLS-TD(0) from $\lambda=0$ to general $0\leq\lambda\leq1$. The convergence with probability 1 of RLS-TD($\lambda$) is proved for ergodic Markov chains and the limit of convergence is also analyzed. In learning prediction experiments for Markov chains, the performance of RLS-TD($\lambda$) and TD($\lambda$) as well as LS-TD($\lambda$) is compared, where a wide range of parameter settings is tested. In addition, the influence of the initialization parameters in RLS-TD($\lambda$) is also discussed. It is observed that the rate of convergence is influenced by the initialization of the variance matrix, which is a phenomenon investigated theoretically in adaptive filtering (Moustakides, 1997; Haykin, 1996).

As will be analyzed in the following sections, there are two benefits of the extension from RLS-TD(0) to RLS-TD($\lambda$). One is that the value of $\lambda$ ($0\leq\lambda\leq1$) will still affect the performance of the RLS-based temporal-difference algorithms. Although for RLS-TD($\lambda$), the rate of convergence is mainly influenced by the initialization of the variance matrix, the bound of approximation error is dominantly determined by the parameter $\lambda$. The smallest error bound can be obtained for $\lambda=1$ and the worst bound is obtained for $\lambda=0$. These bounds suggest that the value of $\lambda$ should be selected appropriately to obtain the best approximation error. The second benefit is that RLS-TD($\lambda$) is more suitable for online learning than LS-TD($\lambda$) since the computation per time-step is reduced from O($K^3$) to O($K^2$), where $K$ is the number of state features.

The Adaptive-Heuristic-Critic (AHC) learning algorithm is a class of reinforcement learning methods that has an actor-critic architecture and can be used to solve full reinforcement learning or learning control problems. By applying the RLS-TD($\lambda$) algorithm in the critic, the Fast-AHC algorithm is proposed in this paper. Using RLS methods in the critic, the performance of learning prediction in the critic is improved so that learning control problems can be solved more efficiently. Simulation experiments on the learning control of the cart-pole balancing problem and the swing-up of an acrobot are conducted to verify the effectiveness of the Fast-AHC method. By comparing with conventional AHC methods which use TD($\lambda$) in the critic, it is demonstrated that Fast-AHC can obtain higher data efficiency than conventional AHC methods. Experiments on the performance comparisons between AHC methods using LS-TD($\lambda$) and Fast-AHC are also conducted. In the learning control experiments, it is also illustrated that the initializing constant of the variance matrix in RLS-TD($\lambda$) influences the performance of Fast-AHC and different values of the constant should be selected to get better performance in different problems. The above results are analyzed based on the theoretical work on the transient phase of RLS methods.

This paper is organized as follows. In Section 2, an introduction on the previous linear temporal-difference algorithms is presented. In Section 3, the RLS-TD($\lambda$) algorithm is proposed and its convergence (with probability one) is proved. In Section 4, a simulation example of the value-function prediction for absorbing Markov chains is presented to illustrate the effectiveness of the RLS-TD($\lambda$) algorithm, where different parameter settings for different algorithms including LS-TD($\lambda$) are studied. In Section 5, the Fast-AHC method is proposed and the simulation experiments on the learning control of the cart-pole balancing and the acrobot are conducted to compare Fast-AHC with the conventional AHC method as well as the LS-TD($\lambda$)-based AHC method. Some simulation results are presented and analyzed in detail. The last section contains concluding remarks and directions for future work.

## 2. Previous Work on Linear Temporal-Difference Algorithms

In this section, a brief discussion on the conventional linear TD($\lambda$) algorithm and RLS-TD(0) as well as the LS-TD($\lambda$) algorithm will be given. First of all, some mathematical notations are presented as follows.

Consider a Markov chain whose states lie in a finite or countable infinite space $S$. The states of the Markov chain can be indexed as $\{1,2,\ldots,n\}$, where $n$ is possibly infinite. Although the algorithms and the results in this paper are applicable to Markov chains with general state space, the discussion in this paper will be restricted within the cases with a countable state space to simplify the notation. The extension to Markov chains with a general state space only requires the translation of the matrix notation into operator notation.

Let the trajectory generated by the Markov chain be denoted by $\{x_t \mid t=0,1,2,\ldots; x_t \in S\}$. The dynamics of the Markov chain is described by a transition probability matrix $P$ whose $(i,j)$-th entry, denoted by $p_{ij}$, is the transition probability for $x_{t+1}=j$ given that $x_t=i$. For each state transition from $x_t$ to $x_{t+1}$, a scalar reward $r_t$ is defined. The value function of each state is defined as follows:

$$V(i) = E\{\sum_{t=0}^{\infty} \gamma^t r_t \big| x_0 = i\} \tag{1}$$

where $0 < \gamma \leq 1$ is a discount factor.

In the TD($\lambda$) algorithm, there are two basic mechanisms which are the temporal difference and the eligibility trace, respectively. Temporal differences are defined as the differences between two successive estimations and have the following form.

$$\delta_t = r_t + \gamma \widetilde{V}_t(x_{t+1}) - \widetilde{V}_t(x_t) \tag{2}$$

where $x_{t+1}$ is the successive state of $x_t$, $\widetilde{V}(x)$ denotes the estimate of the value function $V(x)$ and $r_t$ is the reward received after the state transition from $x_t$ to $x_{t+1}$.

The Eligibility trace can be viewed as an algebraic trick to improve learning efficiency without recording all the data of a multi-step prediction process. This trick is based on the idea of using the truncated return of a Markov chain. In temporal-difference learning with eligibility traces, an $n$-step truncated return is defined as

$$R_t^n = r_t + \gamma r_{t+1} + \dots + \gamma^{n-1} r_{t+n-1} + \gamma^n \widetilde{V}_t(s_{t+n}) \tag{3}$$

For an absorbing Markov chain whose length is $T$, the weighted average of truncated returns is

$$R_t^\lambda = (1-\lambda) \sum_{n=1}^{T-t-1} \lambda^{n-1} R_t^n + \lambda^{T-t-1} R_T \tag{4}$$

where $0 \le \lambda \le 1$ is a decaying factor and $R_T = r_t + \gamma r_{t+1} + \dots + \gamma^T r_T$ is the Monte-Carlo return at the terminal state. In each step of the TD($\lambda$) algorithm, the update rule of the value function estimation is determined by the weighted average of truncated returns defined above. The corresponding update equation is

$$\Delta \widetilde{V}_t(s_i) = \alpha_t (R_t^\lambda - \widetilde{V}_t(s_i)) \tag{5}$$

where $\alpha_t$ is a learning factor.

The update equation (5) can be used only after the whole trajectory of the Markov chain is observed. To realize incremental or online learning, eligibility traces are defined for each state as follows:

$$z_{t+1}(s_i) = \begin{cases} \gamma\lambda\, z_t(s_i) + 1, & \text{if } s_i = s_t \\ \gamma\lambda\, z_t(s_i), & \text{if } s_i \ne s_t \end{cases} \tag{6}$$

The online TD($\lambda$) update rule with eligibility traces is

$$\widetilde{V}_{t+1}(s_i) = \widetilde{V}_t(s_i) + \alpha_t \delta_t z_{t+1}(s_i) \tag{7}$$

where $\delta_t$ is the temporal difference at time step $t$, which is defined in (2) and $z_0(s)=0$ for all $s$.

Since the state space of a Markov chain is usually large or infinite in practice, function approximators such as neural networks are commonly used to approximate the value function. TD($\lambda$) algorithms with linear function approximators are the most popular and well-studied ones.

Consider a general linear function approximator with a fixed basis function vector

$$\phi(x) = (\phi_1(x), \phi_2(x), \dots, \phi_n(x))^T$$

The estimated value function can be denoted as

$$\widetilde{V}_t(x) = \phi^T(x) W_t \tag{8}$$

where $W_t = (w_1, w_2,...,w_n)^T$ is the weight vector.

The corresponding incremental weight update rule is

$$W_{t+1} = W_t + \alpha_t(r_t + \gamma\phi^T(x_{t+1})W_t - \phi^T(x_t)W_t)\vec{z}_{t+1} \tag{9}$$

where the eligibility trace vector $\vec{z}_t(s) = (z_{1t}(s), z_{2t}(s),..., z_{nt}(s))^T$ is defined as

$$\vec{z}_{t+1} = \gamma\lambda\vec{z}_t + \phi(x_t) \tag{10}$$

In Tsitsiklis and Roy (1997), the above linear TD($\lambda$) algorithm is proved to converge with probability 1 under certain assumptions and the limit of convergence $W^*$ is also derived, which satisfies the following equation.

$$E_0[A(X_t)]W^* - E_0[b(X_t)] = 0 \tag{11}$$

where $X_t = (x_t, x_{t+1}, z_{t+1})$ ($t=1,2,...$) form a Markov process, $E_0[\cdot]$ stands for the expectation with respect to the unique invariant distribution of $\{X_t\}$, and $A(X_t)$ and $b(X_t)$ are defined as

$$A(X_t) = \vec{z}_t(\phi^T(x_t) - \gamma\phi^T(x_{t+1})) \tag{12}$$

$$b(X_t) = \vec{z}_t r_t \tag{13}$$

To improve the efficiency of linear TD($\lambda$) algorithms, least-squares methods are used with the linear TD(0) algorithm, and the LS-TD(0) and RLS-TD(0) algorithms are suggested (Brartke and Barto, 1996). In LS-TD(0) and RLS-TD(0), the following quadratic objective function is defined.

$$J = \sum_{t=1}^{T-1}[r_t - (\phi_t^T - \gamma\phi_{t+1}^T)W]^2 \tag{14}$$

Thus, the aim of LS-TD(0) and RLS-TD(0) is to obtain a least-squares estimation of the real value function which satisfies the following Bellman equation.

$$V(x_t) = E[r_t(x_t, x_{t+1}) + \gamma V(x_{t+1})] \tag{15}$$

By employing the instrumental variables approach (Soderstrom and Stoica, 1983), the least-squares solution of (14) is given as

$$W_{LS-TD(0)} = (\sum_{t=1}^{T}(\phi_t(\phi_t - \gamma\phi_{t+1})^T))^{-1}(\sum_{t=1}^{T}\phi_t r_t) \tag{16}$$

where $\phi_t$ is the instrumental variable chosen to be uncorrelated with the input and output noises.

In RLS-TD(0), recursive least-squares methods are used to decrease the computational burden of LS-TD(0). The update rules of RLS-TD(0) are as follows:

$$W_{t+1} = W_t + P_t\phi_t(r_t - (\phi_t - \gamma\phi_{t+1})^T W_t)/(1 + (\phi_t - \gamma\phi_{t+1})^T P_t\phi_t) \tag{17}$$

$$P_{t+1} = P_t - P_t\phi_t(\phi_t - \gamma\phi_{t+1})^T P_t/(1 + (\phi_t - \gamma\phi_{t+1})^T P_t\phi_t) \tag{18}$$

The convergence (with probability one) of LS-TD(0) and RLS-TD(0) is proved for periodic and absorbing Markov chains under certain assumptions (Brartke and Barto,1996).

In Boyan (1999,2002), LS-TD($\lambda$) is proposed by solving (11) directly and the model-based property of LS-TD($\lambda$) is also analyzed. However, for LS-TD($\lambda$), the computation per time-step is O($K^3$), i.e., the cubic order of the state feature number. Therefore the computation required by LS-TD($\lambda$) increases very fast when $K$ increases, which is undesirable for online learning.

In the next section, we propose the RLS-TD($\lambda$) algorithm by making use of recursive least-squares methods so that the computational burden of LS-TD($\lambda$) can be reduced from O($K^3$) to O($K^2$). We also give a rigorous mathematical analysis on the algorithm, where the convergence (with probability 1) of RLS-TD($\lambda$) is proved.

## 3. The RLS-TD($\lambda$) Algorithm

For the Markov chain discussed above, when linear function approximators are used, the least-squares estimation problem of (11) has the following objective function.

$$J = \left\| \sum_{t=1}^{T} A(X_t)W - \sum_{t=1}^{T} b(X_t) \right\|^2 \tag{19}$$

where $A(X_t) \in R^{n \times n}, b(X_t) \in R^n$ are defined as (12) and (13), respectively, $\|\cdot\|$ is a Euclid norm and $n$ is the number of basis functions.

In LS-TD($\lambda$), the least-squares estimate of the weight vector $W$ is computed according to the following equation.

$$W_{LS-TD(\lambda)} = A_T^{-1} b_T = \left( \sum_{t=1}^{T} A(X_t) \right)^{-1} \left( \sum_{t=1}^{T} b(X_t) \right) \tag{20}$$

where

$$A_T = \sum_{t=0}^{T} (A(X_t)) = \sum_{t=0}^{T} \bar{z}_t (\phi^T(x_t) - \gamma \phi^T(x_{t+1})) \tag{21}$$

$$b_T = \sum_{t=0}^{T} b(X_t) = \sum_{t=0}^{T} \bar{z}_t r_t \tag{22}$$

As is well known in system identification, adaptive filtering and control, RLS methods are commonly used to solve the computational and memory problems of least-squares algorithms. In the sequel, we present the RLS-TD($\lambda$) algorithm based on the above idea. First, the matrix inverse lemma is given as follows:

**Lemma 1** (Ljung, et al.,1983). *If* $A \in R^{n \times n}, B \in R^{n \times 1}, C \in R^{1 \times n}$ *and* $A$ *is invertible*, *then*

$$(A + BC)^{-1} = A^{-1} - A^{-1}B(I + CA^{-1}B)^{-1}CA^{-1} \tag{23}$$

Let

$$P_t = A_t^{-1} \tag{24}$$

$$P_0 = \delta I \tag{25}$$

$$K_{t+1} = P_{t+1}\vec{z}_t \tag{26}$$

where $\delta$ is a positive number and $I$ is the identity matrix.

Then the weight update rules of RLS-TD($\lambda$) are given by

$$K_{t+1} = P_t\vec{z}_t /(\mu + (\phi^T(x_t) - \gamma\phi^T(x_{t+1}))P_t\vec{z}_t) \tag{27}$$

$$W_{t+1} = W_t + K_{t+1}(r_t - (\phi^T(x_t) - \gamma\phi^T(x_{t+1}))W_t) \tag{28}$$

$$P_{t+1} = \frac{1}{\mu}[P_t - P_t\vec{z}_t[\mu + (\phi^T(x_t) - \gamma\phi^T(x_{t+1}))P_t\vec{z}_t]^{-1}(\phi^T(x_t) - \gamma\phi^T(x_{t+1}))P_t] \tag{29}$$

where for the standard RLS-TD($\lambda$) algorithm, $\mu$=1; for the general forgetting factor RLS-TD($\lambda$) case, $0<\mu\leq1$.

The forgetting factor $\mu$ ($0<\mu\leq1$) is usually used in adaptive filtering to improve the performance of RLS methods in non-stationary environments. The forgetting factor RLS-TD($\lambda$) algorithm with $0<\mu\leq1$ can be derived using similar techniques as in Haykin (1996). The detailed derivation of RLS-TD($\lambda$) is referred to Appendix A.

In the follows, the descriptions of RLS-TD($\lambda$) for two different kinds of Markov chains are given. First, a complete description of RLS-TD($\lambda$) for ergodic Markov chains is presented below.

---

**Algorithm 1** RLS-TD($\lambda$) for ergodic Markov chains

---

1: **Given:**
   - A termination criterion for the algorithm.
   - A set of basis functions $\{\phi_j(i)\}$ ($j$=1,2,…,$n$) for each state $i$, where $n$ is the number of basis functions.

2: **Initialize:**
   (2.1) Let $t$=0.
   (2.2) Initialize the weight vector $W_t$, the variance matrix $P_t$, the initial state $x_0$.
   (2.3) Set the eligibility traces vector $\vec{z}_0$ =0.

3: **Loop:**
   (3.1) For the current state $x_t$, observe the state transition from $x_t$ to $x_{t+1}$ and the reward $r(x_t, x_{t+1})$.
   (3.2) Apply equations (27)-(29) to update the weight vector.
   (3.3) $t$=$t$+1.
   **until the termination criterion is satisfied**.

---

The RLS-TD($\lambda$) algorithm for absorbing Markov chains is a little different from the above algorithm in coping with the state features of absorbing states. Following is a description of RLS-TD($\lambda$) for absorbing Markov chains.

---

**Algorithm 2** RLS-TD($\lambda$) for absorbing Markov chains

---

1: **Given:**
- A termination criterion for the algorithm.
- A set of basis functions $\{\phi_j(i)\}$ ($j$=1,2,…,$n$) for each state $i$, where $n$ is the number of basis functions.

2: **Initialize:**

(2.1) Let $t$=0.

(2.2) Initialize the weight vector $W_t$, the variance matrix $P_t$, the initial state $x_0$.

(2.3) Set the eligibility traces vector $\vec{z}_0$=0.

3: **Loop:**

(3.1) For the current state $x_t$,
- **If** $x_t$ is an absorbing state, set $\phi(x_{t+1})$=0, $r(x_t)$=$r_T$, where $r_T$ is the terminal reward.
- **Otherwise**, observe the state transition from $x_t$ to $x_{t+1}$ and the reward $r(x_t, x_{t+1})$.

(3.2) Apply equations (27)-(29) to update the weight vector.

(3.3) **If** $x_t$ is an absorbing state, re-initialize the process by setting $x_{t+1}$ to an initial state and set the eligibility traces $\vec{z}_t$ to a zero vector.

(3.4) $t$=$t$+1.

**until the termination criterion is satisfied**.

---

In the above RLS-TD($\lambda$) algorithm for absorbing Markov chains, the weight updates in the absorbing states are treated differently and the process is re-initialized in absorbing states to transform the absorbing Markov chain into an equivalent ergodic Markov chain. So in the following convergence analysis, we only focus on ergodic Markov chains.

Under similar assumptions as in Tsitsiklis and Roy (1997), we will prove that the proposed RLS-TD($\lambda$) algorithm converges with probability one.

**Assumption 1.** *The Markov chain $\{x_t\}$, whose transition probability matrix is P, is ergodic, and there is a unique distribution $\pi$ that satisfies*

$$\pi^T P = \pi^T \qquad (30)$$

*with $\pi(i)$>0 for all $i \in S$ and $\pi$ is a finite or infinite vector, depending on the cardinality of S.*

**Assumption 2.** *Transition rewards $r(x_t, x_{t+1})$ satisfy*

$$E_0[r^2(x_t, x_{t+1})] < \infty \qquad (31)$$

*where $E_0[\ ]$ is the expectation with respect to the distribution $\pi$.*

**Assumption 3.** *The matrix $\Phi = [\phi_1, \phi_2, ..., \phi_n] \in R^{N \times n}$ has full column rank, that is, the basis*

*functions* $\phi_i$ *(i=1,2,…,n) are linearly independent.*

**Assumption 4.** *For every i (i=1,2,…,n), the basis function* $\phi_i$ *satisfies*

$$E_0[\phi_i^2(x_t)] < \infty \tag{32}$$

**Assumption 5.** *The matrix* $[P_0^{-1} + \frac{1}{T}\sum_{t=1}^{T} A(X_t)]$ *is non-singular for all T>0.*

Assumptions 1–4 are almost the same as those for the linear TD($\lambda$) algorithms discussed in Tsitsiklis and Roy (1997) except that in Assumption 1, ergodic Markov chains are considered. Assumption 5 is specially needed for the convergence of the RLS-TD($\lambda$) algorithm.

Based on the above assumptions, the convergence theorem for RLS-TD($\lambda$) can be given as follows:

**Theorem 1.** *For a Markov chain which satisfies Assumptions 1–5, the asymptotic estimate found by RLS-TD($\lambda$) converges, with probability 1, to W\* determined by* (11).

For the proof of Theorem 1, please refer to Appendix B. The condition specified by Assumption 5 can be satisfied by setting $P_0 = \delta I$ appropriately.

According to Theorem 1, RLS-TD($\lambda$) converges to the same solution as conventional linear TD($\lambda$) algorithms do, which satisfies (11). So the limit of convergence can be characterized by the following theorem.

**Theorem 2** (Tsitsiklis and Roy ,1997) *Let W\* be the weight vector determined by* (11) *and V\* be the true value function of the Markov chain, then under Assumption 1–4, the following relation holds.*

$$\left\| \Phi W^* - V^* \right\|_D \le \frac{1-\lambda\gamma}{1-\gamma} \left\| \Pi V^* - V^* \right\|_D \tag{33}$$

*where* $\left\| X \right\|_D = \sqrt{X^T D X}$ , $\Pi = \Phi(\Phi^T D \Phi)^{-1} \Phi^T D$ .

For more explanations on the notations in Theorem 2, please refer to Appendix B.

As discussed by Tsitsiklis and Roy (1997), the above theorem shows that the distance of the limiting function $\Phi W^*$ from the true value function $V^*$ is bounded and the smallest bound of approximation error can be obtained when $\lambda$=1. For every $\lambda$<1, the bound actually deteriorates as $\lambda$ decreases. The worst bound is obtained when $\lambda$=0. Although this is only a bound, it strongly suggests that higher values of $\lambda$ are likely to produce more accurate approximations of $V^*$.

Compared to LS-TD($\lambda$), there is an additional parameter in RLS-TD($\lambda$), which is the value $\delta$ for the initial variance matrix $P_0$. As was pointed out by Haykin (1996,pp.570), the exact value of the initializing constant $\delta$ has an insignificant effect when the data length is large enough. This means that in the limit, the final solutions obtained by LS and RLS are almost the same. For the influence of $\delta$ on the transient phase, when the positive constant $\delta$ becomes large enough or goes to infinity, the transient behavior of RLS will be almost the same as that of LS methods (Ljung, 1983). But when $\delta$ is initialized with a relatively small value, the transient phases of RLS and LS will be different. In practice, it is observed that there is a variable performance of RLS as a function of the initialization of $\delta$ (Moustakides, 1997). In some cases, RLS can exhibit a significantly faster convergence when initialized with a relatively small positive definite matrix than when initialized with a large one (Haykin,1996; Moustakides, 1997; Hubing and Alexander, 1989). A first effort toward this direction is the statistical analysis of RLS for soft and exact initialization but limits to the case that the number of iterations is less than the size of the estimation vector (Hubing and Alexander, 1989). Moustakides (1997) provided a theoretical analysis on the relation between the algorithmic performance of RLS and the initialization of $\delta$. By using the settling time as the performance measure, Moustakides proved that the well-known rule of initialization with a relatively small matrix is preferable for cases of high and medium signal-to-noise ratio (SNR), whereas for low SNR, a relatively large matrix must be selected for achieving best results. In the following learning prediction experiments of RLS-TD($\lambda$), as well as the learning control simulation of Fast-AHC, it is observed that the value of the initializing constant $\delta$ also plays an important role in the convergence performance, and the above theoretical analyses provide a clue to explain our experimental results.

## 4. Learning Prediction Experiments on Markov Chains

In this section, an illustrative example is given to show the effectiveness of the proposed RLS-TD($\lambda$) algorithm. Furthermore, the algorithmic performance under the influence of the initializing constant $\delta$ is studied.

The example is a finite-state absorbing Markov chain called the Hop-World problem (Boyan, 1999). As shown in Figure 1, the Hop-World problem is a 13-state Markov chain with an absorbing state.

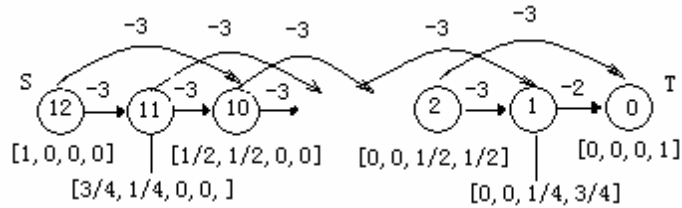

Figure 1: The Hop-World Problem

In Figure 1, state 12 is the initial state for each trajectory and state 0 is the absorbing state. Each non-absorbing state has two possible state transitions with transition probability 0.5. Each state transition has reward –3 except the transition from state 1 to state 0 which has a reward of –2. Thus, the true value function for state $i$ ($0 \leqslant i \leqslant 12$) is $-2i$.

To apply linear temporal-difference algorithms to the value function prediction problem, a set of four-element state features or basis functions is chosen, as shown in Figure 1. The state features of states 12,8,4 and 0 are, respectively, [1,0,0,0], [0,1,0,0], [0,0,1,0], [0,0,0,1] and the state features of other states are obtained by linearly interpolating between these.

In our simulation, the RLS-TD($\lambda$) algorithm as well as LS-TD($\lambda$) and conventional linear TD($\lambda$) algorithms are used to solve the above value function prediction problem without knowing the model of the Markov chain. In the experiments, a trial is defined as the period from the initial state 12 to the terminal state 0. The performance of the algorithms is evaluated by the averaged root mean squared (RMS) error of value-function predictions over all the 13 states. For each parameter setting, the performance is averaged over 20 independent Monte-Carlo runs. Figure 2 shows the learning curves of RLS-TD($\lambda$) and conventional linear TD($\lambda$) algorithms with three different parameter settings. The parameter $\lambda$ is set to 0.3 for all the algorithms and the step-size parameter of TD($\lambda$) has the following form.

$$\alpha_n = \alpha_0 \frac{N_0 + 1}{N_0 + n} \tag{34}$$

The above step-size schedule is also studied in Boyan (1999). In our experiments, three different settings are used, which are

(s1)  $\alpha_0 = 0.01$, $N_0 = 10^6$

(s2)  $\alpha_0 = 0.01$, $N_0 = 1000$  (35)

(s3)  $\alpha_0 = 0.1$, $N_0 = 1000$.

Different from those in Boyan (1999), the linear TD($\lambda$) algorithms applied here are in their online forms, which update the weights after every state transitions. So the parameter $n$ in (34) is the number of state transitions. In each run, the weights are all initialized to zeroes. In Figure 2, the learning curves of conventional linear TD($\lambda$) algorithms with step-size schedules (s1), (s2) and (s3) are shown by curves 1,2 and 3, respectively. For each curve, the averaged RMS errors of value function predictions over all the states and 20 independent runs are plotted for each trial. Curve 4 shows the learning performance of RLS-TD($\lambda$). One additional parameter for RLS-TD($\lambda$) is the initial value $\delta$ of the variance matrix $P_0$. In this experiment, $\delta$ is set to 500, which is a relatively large value. From Figure 2, it can be concluded that by making use of RLS methods, RLS-TD($\lambda$) can obtain much better performance than conventional linear TD($\lambda$) algorithms and eliminates the design problem of the step-size schedules. Other experiments for linear TD($\lambda$) and RLS-TD($\lambda$) with different parameters $\lambda$ are also conducted and similar results are obtained when the initial values $\delta$ of RLS-TD($\lambda$) are large and the conclusion is confirmed.

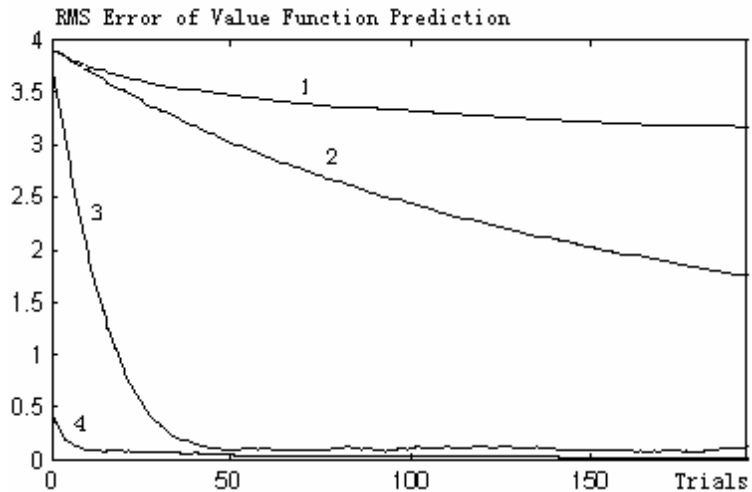

Figure 2: Performance comparison between RLS-TD($\lambda$) and TD($\lambda$)

1,2,3 ---TD(0.3) with step-size parameters specified by (s1),(s2) and (s3)

4—RLS-TD(0.3) with initial variance matrix $P_0 = 500I$

We have done demonstrative experiments to investigate the influence of $\delta$ on the performance of the RLS-TD($\lambda$) algorithm. Figure 3 shows the performance comparison between RLS-TD($\lambda$) algorithms using two different initial parameters of the variance matrix $P_0$, which are $P_0$=0.1$I$ and $P_0$=1000$I$, respectively. The forgetting factor is $\mu$=0.995. The performance of the suggested algorithm is measured by the averaged RMS errors of the value function prediction in the first 200 trials over 20 independent runs and all the 13 states. In the experiments, 11 settings of the parameter $\lambda$ are tested, which are 0.1$n$ ($n$=0,1,…,10).

In Figure 3, it is clearly shown that the performance of RLS-TD($\lambda$) with a large initial value of $\delta$ is much better than RLS-TD($\lambda$) with a small initial value of $\delta$. In other experiments with different parameter settings of $\lambda$ and $\delta$, similar results are also obtained. We may refer this phenomenon to the low SNR case of the forgetting factor RLS studied in Moustakides (1997). For the Hop-World problem, the stochastic state transitions could introduce high equation residuals $A(X_t)W - b(X_t)$ in (19), which corresponds to the additive noise with large variance, i.e., the low SNR case. As has been discussed in Section 2, for the forgetting factor RLS in low SNR cases, a relatively large initializing constant $\delta$ must be selected for better results. A full understanding of this phenomenon is yet to be found.

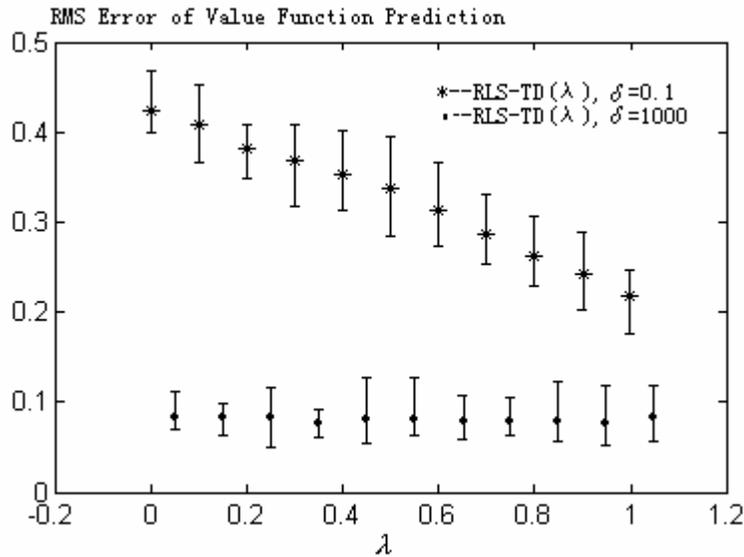

Figure 3: Performance comparison of RLS-TD($\lambda$) with different initial value of $\delta$ ($\mu$=0.995)

The performance of RLS-TD($\lambda$) with unit forgetting factor $\mu$=1 is also tested in our experiments. Although the initial value effect in RLS with $\mu$=1 has not been discussed intensively (Moustakides,1997), the same effects of $\delta$ are observed empirically in the case of $\mu$=1 as that in $\mu$<1, which is shown by Figure 4.

In our other experiments, it is also found that when $\delta$ is initialized with a small value, the performance is sensitive to the values of $\delta$ and the parameter $\lambda$. In this case, the convergence speed of RLS-TD($\lambda$) increases as $\lambda$ increases from 0 to 1, which is shown in Figure 3. Furthermore, when $\lambda$ is fixed, the performance of RLS-TD($\lambda$) deteriorates as $\delta$ becomes smaller, as shown in Figure 5 .

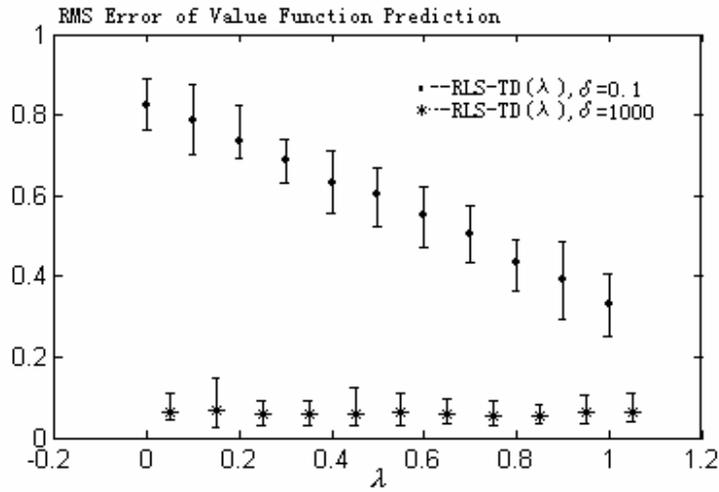

Figure 4: Performance comparison of RLS-TD(λ) with different initial value of δ (μ=1)

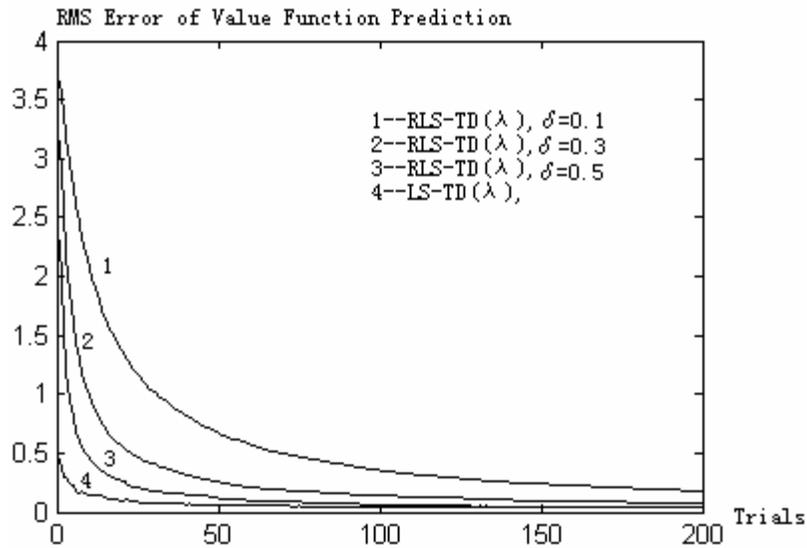

Figure 5: Learning curves of LS-TD(λ) and RLS-TD(λ) with different δ (μ=1)

In Figure 5, the learning curves of RLS-TD(λ) with different initializing constants δ are shown and compared with that of LS-TD(λ). In the experiment, λ is set to 0.5. From Figure 5, it is shown that the performance of RLS-TD(λ) approaches that of LS-TD(λ) when δ becomes large. As is well known, when δ becomes large enough, the performance of RLS and LS methods will be almost the same. Figure 6 shows the performance comparison between LS-TD(λ) and RLS-TD(λ) with a large value of δ. The initial variance matrix for RLS-TD(λ) is set to 500$I$ in every runs, where $I$ is the identity matrix.

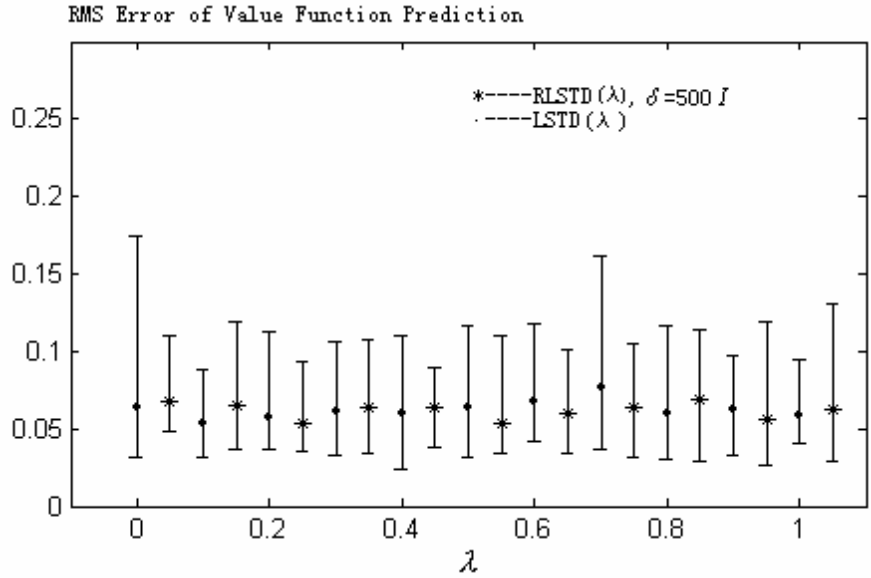

Figure 6: Performance comparison of LS-TD(λ) and RLS-TD(λ) with $\mu$=1 and large initial value of $\delta$

Based on the above experimental results, it can be concluded that the convergence speed of RLS-TD($\lambda$) is mainly influenced by the initial value $\delta$ of the variance matrix and the parameter $\lambda$. Detailed discussions on the properties of RLS-TD($\lambda$) are given as follows:

(1)   When $\delta$ is relatively large, the effect of $\lambda$ becomes small. If $\delta$ is large enough or goes to infinity, the performance of RLS-TD($\lambda$) and LS-TD($\lambda$) will be almost the same, as was discussed above. In such cases, the effect of $\lambda$ on the speed of convergence is insignificant, which coincides with the discussion in Boyan (1999). However, as described in Theorem 2, the value of $\lambda$  still affects the ultimate error bound of value function approximation.

(2)   When $\delta$ is relatively small, it is observed that the convergence performance of RLS-TD(λ) is different from that of LS-TD(λ) and is influenced by the values of both $\delta$ and λ. In the experiments of the Hop-World problem, the results show that smaller values of $\delta$ lead to slower convergence. These results may be explained by the theoretical analysis on the transient phase of the forgetting factor RLS (Moustakides,1997). According to the theory in Moustakides (1997), larger values of $\delta$ are needed for better performance in the cases of low SNR while smaller $\delta$ values are preferable for fast convergence in the cases of high and medium SNR. So different values of $\delta$ must be selected for faster convergence of RLS-TD($\lambda$) in different cases. Especially, in some cases, such as the high SNR case discussed in Moustakides (1997), RLS methods with small values of $\delta$ can obtain a very fast speed of convergence.

(3)   Compared to conventional linear TD($\lambda$) algorithms, the RLS-TD($\lambda$) algorithm can obtain much better performance by making use of RLS methods for value function prediction problems. Furthermore, in TD($\lambda$), a step-size schedule needs to be carefully designed to achieve good performance, while in RLS-TD($\lambda$), the initial value $\delta$ of the variance matrix can be selected according to the criterion of a "large" or a "small" value.

(4)   For the comparison of LS-TD($\lambda$) and RLS-TD($\lambda$), which one is preferable depends on the objective. In online applications, RLS-TD($\lambda$) has advantages in computational efficiency because the computation per step for RLS-TD($\lambda$) is O($K^2$) and for LS-TD($\lambda$), it is O($K^3$), where $K$ is the number of state features. Moreover, as will be seen later, RLS-TD($\lambda$) can obtain better transient convergence performance than LS-TD($\lambda$) in some cases. On the other hand, LS-TD($\lambda$) may be preferable to RLS-TD($\lambda$) in the long-term convergence performance, as can be seen in Figure 5. And from a system identification point of view, LS-TD($\lambda$) can obtain unbiased parameter estimates in face of white additive noises while RLS-TD($\lambda$) with finite $\delta$ would possess large parameter discrepancies.

## 5. The Fast-AHC Algorithm and Two Learning Control Experiments

In this section, the Fast-AHC algorithm is proposed based on the above results on learning prediction to solve learning control problems. Two learning control experiments are conducted to illustrate the efficiency of Fast-AHC.

### 5.1 The Fast-AHC Algorithm

The ultimate goal of reinforcement learning is learning control, i.e., to estimate the optimal policies or the optimal value functions of Markov decision processes (MDPs). Until now, several reinforcement learning control algorithms including Q-learning (Watkins and Dayan,1992), Sarsa-learning (Singh, et al.,2000) and the Adaptive Heuristic Critic (AHC) algorithm (Barto, Sutton and Anderson,1983) have been proposed. Among the above methods, the AHC method is different from Q-learning and Sarsa-learning which are value-function-based methods. In the AHC method, value functions and policies are separately represented while in value-function-based methods the policies are determined by the value functions directly. There are two components in the AHC method, which are called the critic and the actor, respectively. The actor is used to generate control actions according to the policies. The critic is used to evaluate the policies represented by the actor and provide the actor with internal rewards without waiting for delayed external rewards. Since the objective of the critic is policy evaluation or learning prediction, temporal-difference learning methods are chosen as the critic's learning algorithms. The learning algorithm of the actor is determined by the estimation of the gradient of the policies. In the following discussion, a detailed introduction on the AHC method is given.

Figure 7 shows the architecture of a learning system based on the AHC method. The learning system consists of a critic network and an actor network. The inputs of the critic network include the external rewards and the state feedback from the environment. The internal rewards provided by the critic network are called the temporal-difference (TD) signals.

As in most reinforcement learning methods, the whole system is modeled as an MDP denoted by a tuple $\{S,A,P,R\}$,where $S$ is the state set, $A$ is the action set, $P$ is the state transition probability and $R$ is the reward function. The policy of the MDP is defined as a function $\pi :S \rightarrow \Pr(A)$, where $\Pr(A)$ is a probability distribution in the action space. The objective of the AHC method is to estimate the optimal policy $\pi^*$ satisfying the following equation.

$$J^* = \max_\pi J_\pi = \max_\pi E_\pi [\sum_{t=0}^\infty \gamma^t r_t] \tag{36}$$

where $\gamma$ is the discount factor and $r_t$ is the reward at time-step $t$, $E_\pi[\ ]$ stands for the expectation with respect to the policy $\pi$ and the state transition probabilities and $J_\pi$ is the expected total reward.

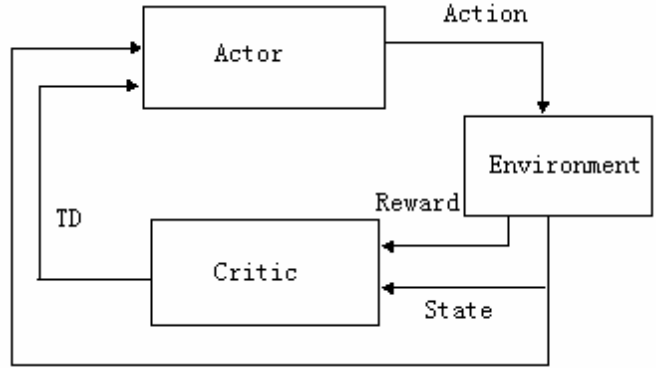

Figure 7: The AHC learning system

The value function for a stationary policy $\pi$ and the optimal value function for the optimal policy are defined as follows:

$$V^{\pi}(s) = E_{\pi}[\sum_{t=0}^{\infty} \gamma^{t} r_{t} | s_{0} = s] \tag{37}$$

$$V^{*}(s) = E_{\pi*}[\sum_{t=0}^{\infty} \gamma^{t} r_{t} | s_{0} = s] \tag{38}$$

According to the theory of dynamic programming, the optimal value function satisfies the following Bellman equation.

$$V^{*}(s) = \max_{a}[R(s,a) + \gamma E V^{*}(s')] \tag{39}$$

where $R(s,a)$ is the expected reward received after taking action $a$ in state $s$.

In AHC, the critic uses temporal-difference learning to approximate the value function of the current policy. When linear function approximators are used in the critic, the weight update equation is

$$W_{t+1} = W_{t} + \alpha_{t}[r_{t} + \gamma V(s_{t+1}) - V(s_{t})]z_{t} \tag{40}$$

where $z_{t}$ is the eligibility trace defined in (10).

The action selection policy of the actor is determined by the current state and the value function estimation of the critic. Suppose a neural network with weight vector $u=[u_{1}, u_{2},\ldots, u_{m}]$ is used in the actor, and the output of the actor network is

$$\bar{y}_{t} = f(u,s_{t}) \tag{41}$$

The action outputs of the actor are determined by the following Gaussian probabilistic distribution.

$$p_{r}(y_{t}) = \exp(-\frac{(y_{t} - \bar{y}_{t})^{2}}{\sigma_{t}^{2}}) \tag{42}$$

where the mean value is given by (41) and the variance is given by

$$\sigma_{t} = k_{1}/(1 + \exp(k_{2} V(s_{t}))) \tag{43}$$

In the above equation, $k_{1}$ and $k_{2}$ are positive constants and $V(s_{t})$ is the value function es-timation of the critic network.

To obtain the learning rule of the actor, an estimation of the policy gradient is given as follows:

$$\frac{\partial J_\pi}{\partial u} = \frac{\partial J_\pi}{\partial \overline{y}_t} \frac{\partial \overline{y}_t}{\partial u} \approx \hat{r}_t \frac{y_t - \overline{y}_t}{\sigma_t} \frac{\partial \overline{y}_t}{\partial u} \tag{44}$$

where $\hat{r}_t$ is the internal reward or the TD signal provided by the critic:

$$\hat{r}_t = r_t + \gamma V(s_{t+1}) - V(s_t) \tag{45}$$

Since in the AHC method, the critic is used to estimate the value function of the actor's policy and provide the internal reinforcement using temporal-difference learning algorithms, the efficiency of temporal-different learning or learning prediction will greatly influence the whole learning system's performance. Although the policy of the actor is changing, it may change relatively slowly especially when fast convergence of learning prediction in the critic can be realized. In the previous sections, RLS-TD($\lambda$) is shown to have better data efficiency than conventional linear TD($\lambda$) algorithms and a very fast convergence speed can be obtained when the initializing constant is chosen appropriately. Thus, applying RLS-TD($\lambda$) to the policy evaluation in the critic network will improve the learning prediction performance of the critic and is promising to enhance the whole system's learning control performance. Based on the above idea, a new AHC method called the Fast-AHC algorithm is proposed in this paper. The efficiency of the Fast-AHC algorithm is verified empirically and detailed analysis of the results is given. Following is a complete description of the Fast-AHC algorithm.

---

**Algorithm 3:** The Fast-AHC algorithm

---

1: **Given:** a critic neural network and an actor neural network, which are both linear in parameters, a stop criterion for the algorithm.

2: **Initialize** the state of the MDP and the learning parameters, set $t$=0.

3: **While** the stop criterion is not satisfied,

    (3.1)    According to the current state $s_t$, compute the output of the actor network $\overline{y}_t$, determine the actual action of the actor by the probability distribution given by (42).

    (3.2)    Take the action $y_t$ on the MDP, and observe the state transition from $s_t$ to $s_{t+1}$, set reward $r_t = r(s_t, s_{t+1})$.

    (3.3)    Apply the RLS-TD($\lambda$) algorithm described in (27)-(29) to update the weights of the critic network.

    (3.4)    Apply the following equation to update the weights of the actor network,

$$a_{t+1} = a_t + \beta_t \frac{\partial J_\pi}{\partial a_t} \tag{46}$$

        where $\beta_t$ is the learning factor of the actor.

    (3.5)    Let $t$=$t$+1, return to 3.

---

**5.2 Learning Control Experiments on The Cart-Pole Balancing Problem**

The balancing control of inverted pendulums is a typical nonlinear control problem and has been widely studied not only in control theory but also in artificial intelligence. In the research of artificial intelligence, the learning control of inverted pendulums is considered as a standard test problem for machine learning methods, especially for RL algorithms. It has been studied in the early work of Michie's BOXES system (Michie,et al.,1968) and later in Barto and Sutton (1983), where the learning controllers only have two output values: +10($N$) and –10($N$). In Berenji, et al.(1992) and Lin, et al.(1994), AHC methods with continuous outputs are applied to the cart-pole balancing problem. In this paper, the cart-pole balancing problem with continuous control values is used to illustrate the effectiveness of the Fast-AHC method.

Figure 8 shows a typical cart-pole balancing control system, which consists of a cart moving horizontally and a pole with one end fixed at the cart. Let $x$ denote the horizontal distance between the center of the cart and the center of the track, where $x$ is negative when the cart is in the left part of the track. Variable $\theta$ denotes the angle of the pole from its upright position (in degrees) and $F$ is the amount of force ($N$) applied to the cart to move it towards its left or right. So the control system has four state variables $x, \dot{x}, \theta, \dot{\theta}$, where $\dot{x}, \dot{\theta}$ are the derivatives of $x$ and $\theta$, respectively.

In Figure 8, the mass of the cart is $M$=1.0kg, the mass of the pole is $m$=0.1kg, the half-pole length is $l$=0.5m, the coefficient of friction of the cart on the track is $\mu_c$=0.0005 and the coefficient of friction of the pole on the cart is $\mu_p$=0.000002. The boundary constraints on the state variables are given as follows.

$$-12° \leq \theta \leq 12° \tag{47}$$
$$-2.4m \leq x \leq 2.4m \tag{48}$$

The dynamics of the control system can be described by the following equations.

$$\begin{cases} \ddot{\theta} = \dfrac{(m+M)g\sin\theta - \cos\theta[F + ml\dot{\theta}^2\sin\theta - \mu_c \, \mathrm{sgn}(\dot{x})] - \dfrac{\mu_p(m+M)\dot{\theta}}{ml}}{\dfrac{4}{3}(M+m)l - ml\cos^2\theta} \\[4mm] \ddot{x} = \dfrac{F + ml(\dot{\theta}^2\sin\theta - \ddot{\theta}\cos\theta) - \mu_c \, \mathrm{sgn}(\dot{x})}{M+m} \end{cases} \tag{49}$$

where $g$ is the acceleration due to the gravity, which is –9.8$m/s^2$. The above parameters and dynamics equations are the same as those studied in Barto et al. (1983).

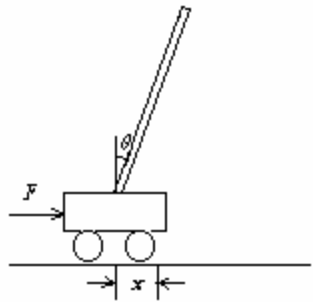

Figure 8: The cart-pole balancing control system

In the learning control experiments of the pole-balancing problem, the dynamics (49) is assumed to be unknown to the learning controller. In addition to the four state variables, the only available feedback is a failure signal that notifies the controller when a failure occurs, which means the values of the state variables exceed the boundary constraints prescribed by inequalities (47) and (48). It is a typical reinforcement learning problem, where the failure signal serves as the reward. Since an external reward may only be available after a long sequence of actions, the critic in the AHC learning controller is used to provide the internal reinforcement signal to accomplish the learning task. Learning control experiments on the pole-balancing problem are conducted using conventional AHC method which uses linear TD($\lambda$) algorithms in the critic and the Fast-AHC method proposed in this paper.

To solve the continuous state space problem in reinforcement learning, a class of linear function approximators, which is called Cerebellar Model Articulation Controller (CMAC) is used. As a neural network model based on the neuro-physiological theory about human cerebellar，CMAC was first proposed by Albus (1975) and has been widely used in automatic control and function approximation. In CMAC neural networks, the dependence of adjustable parameters or weights with respect to outputs is linear. For detailed discussion on the structure of CMAC neural networks, one may refer to Albus (1975) and Sutton & Barto (1998).

In the AHC and Fast-AHC learning controllers, two CMAC neural networks with four inputs and one output for each are used as the function approximators in the critic and the actor, respectively. Each CMAC has $C$ tilings and $M$ partitions for every input. So the total physical memory for each CMAC network is $M^4C$. To reduce the computation and memory requirements, a hashing technique described by the following equations is employed in our experiments. (For detailed discussion on the parameters of the CMAC networks, please refer to Appendix C).

$$A(s) = \sum_{i=1}^{4} [a(i) + M^{i-1}] \tag{50}$$

$$F(s) = A(s) \bmod K \tag{51}$$

In (50) and (51), $s$ represents an input state vector, $a(i)$ ($0 \le a(i) \le M$) is the activated tile for the $i$-th element of $s$, $K$ is the total number of the physical memory and $F(s)$ is the physical memory address corresponding to the state $s$, which is the remainder of $A(s)$ divided by $K$.

In order to compare the performance of different learning algorithms, the initial parameters of each learning controller are selected as follows: The weights of the critic are all initialized to 0 and the weights of the actor are initialized to random numbers in interval [0,0.1]. The other parameters for the AHC and Fast-AHC algorithms are $\gamma = 0.95$, $k_1 = 0.4$ and $k_2 = 0.5$.

In all the experiments, a trial is defined as the period from an initial state to a failure state and the initial state of each trial is set to a randomly generated state near the unstable equilibrium (0,0,0,0) with a maximum distance of 0.05. Equation (49) is employed to simulate the dynamics of the system using the Euler method, which has a time step of 0.02s. When a trial lasts for more than 120,000 time steps, it is said to be successful and the learning controller is assumed to be able to balance the pole. The reinforcement signal for the problem is defined as

$$r_t = \begin{cases} -1, & \text{if failure occurs} \\ 0, & \text{otherwise} \end{cases} \tag{52}$$

The performance of the Fast-AHC method is tested extensively, where different parameter settings including $\lambda$ and the initial variance matrix $P_0$ are chosen. In the experiments, the forgetting factor of RLS-TD(λ) in the critic is set to a value that is equal to 1 or very close to 1. The learning control experiments using conventional AHC methods are also conducted for comparison. The performance comparisons between the two algorithms are shown in Figure 9, 10 and 11.

In the above experiments, the initial variance matrixes of the Fast-AHC algorithm are all set to $P_0 = 0.1I$. The performance of Fast-AHC is compared with AHC for different λ. The numbers of physical memories of the critic network and the actor network are chosen as 30 and 80, respectively. For each parameter setting of the two algorithms, 5 independent runs are tested. The performance is evaluated according to the trial number needed to successfully balance the pole. The learning factors for the actor networks are all set to 0.5, which is a manually optimized value for both algorithms. In all the experiments, 11 settings of λ are tested.

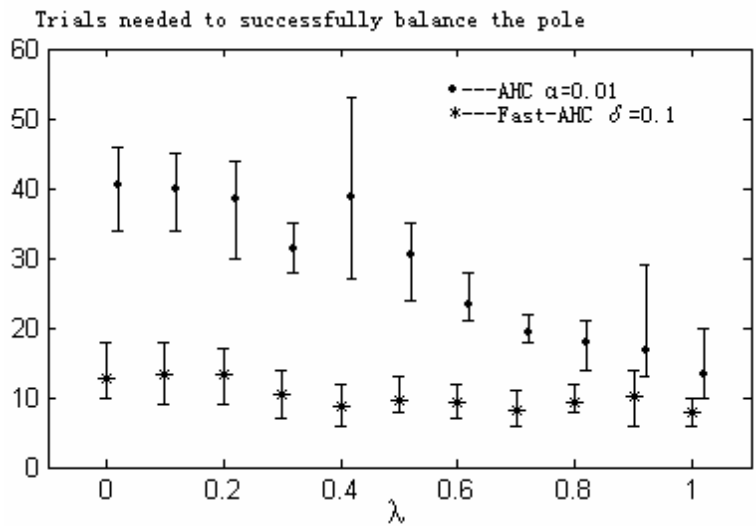

Figure 9: Performance comparison between Fast-AHC and AHC with α=0.01

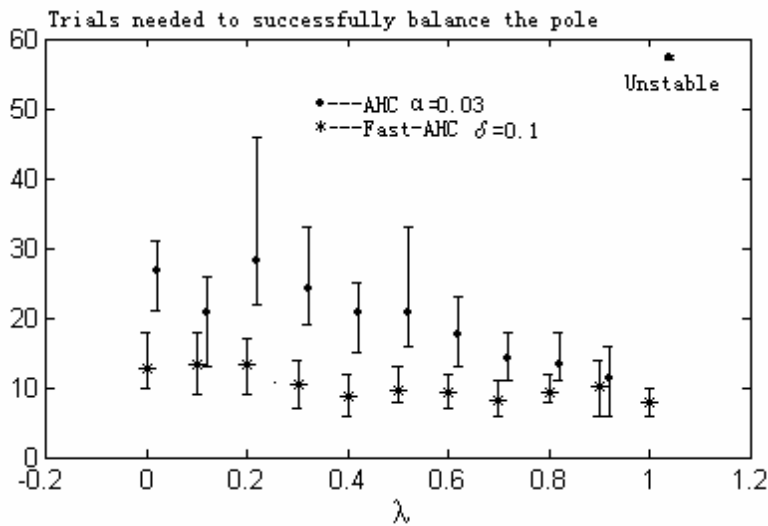

Figure 10: Performance comparison between Fast-AHC and AHC with α=0.03

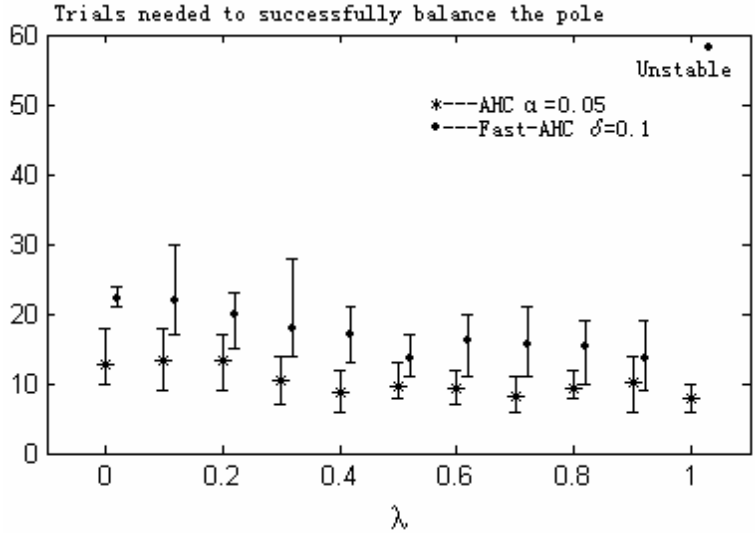

Figure 11: Performance comparison between Fast-AHC and AHC with α=0.05

In Figure 9, 10 and 11, the learning factors of the critic networks in AHC are chosen as α=0.01, 0.03 and 0.05, respectively. It is found that when α<0.01, the performance of AHC becomes worse. For the learning factors that are greater than 0.05, the AHC algorithm may become unstable, and even when α=0.03 and α=0.05, the AHC algorithm becomes unstable for λ=1. For the time-varying learning factors specified in (s1)-(s3), the performance is worse than the above constant learning factors. So the above three settings of the learning factor α are typical and near optimal for the AHC algorithm.

From the above experimental results, it can be concluded that by using RLS-TD(λ) in the critic network, the Fast-AHC algorithm can obtain better performance than conventional AHC algorithms. Although Fast-AHC requires more computation per step than AHC, it is more efficient than AHC in that less trials or data are needed to successfully balance the pole.

As has been discussed in the previous sections, the convergence performance of RLS-TD(λ) is influenced by the initial value of the variance matrix. This is also the case in Fast-AHC. In the above learning control experiments, a small value δ=0.1 is selected. In other experiments, when δ is set to other small values, the performance of Fast-AHC is satisfactory and is better than AHC. However, when δ is equal to a relatively large value, for example δ=100 or 500, the performance of Fast-AHC deteriorates significantly. Since RLS-TD(λ) with a large initializing constant has similar performance as LS-TD(λ), it can be deduced that the AHC method using LS-TD(λ) in the critic will also have bad performance in the cart-pole balancing problem. To verify this, experiments are conducted using Fast-AHC with large initializing constant δ and AHC using LS-TD(λ). For each parameter setting, 5 independent runs are tested. In the experiments, the maximum trials for each algorithm in one run is 200 so that if an algorithm fails to balance the pole within 200 trials, its performance is set to 200.When using LS-TD(λ) in the AHC method, there may be computational problems in the matrix inversion during the first few steps of learning and two methods are tried to avoid this problem. One is the usage of TD(λ) in the first 60 steps of updates. The other is that the actor is not updated in the early stage of learning until LS-TD(λ) is stable. However, similar results are found for the two methods. Figure 12 shows the experimental results which clearly verify that the performance of Fast-AHC with a large initializing constant $\delta$ is similar to AHC using LS-TD($\lambda$) and it is much worse than Fast-AHC with a small $\delta$. A detailed discussion of this phenomenon is provided in subsection 5.4.

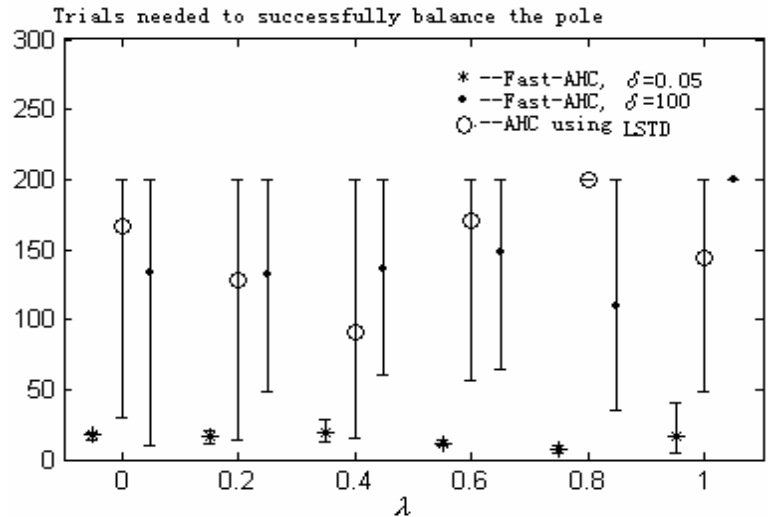

Figure 12: Performance comparison of Fast-AHC with different initial variance

In the following Figure 13 and Figure 14, the variations of the pole angle $\theta$ and the control force $F$ are plotted, where a successfully trained Fast-AHC learning controller is used to control the cart-pole system.

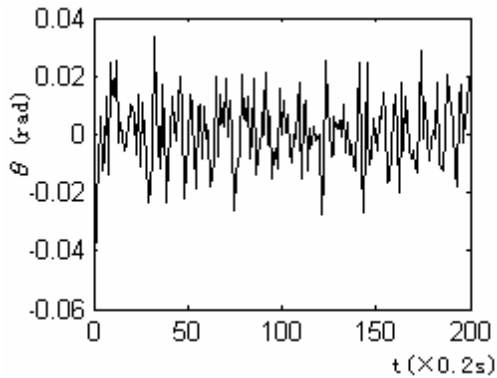
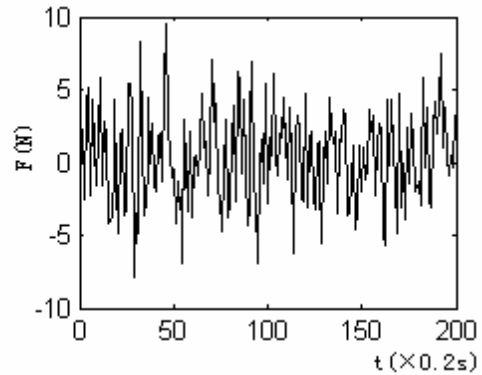

Figure 13: Variation of the pole angle          Figure 14: Variation of the control force

## 5.3 Learning Control Experiments of The Acrobot

In this subsection, another learning control example, which is the swing-up control of the acrobot in minimum time, is presented. The learning control of the acrobot is a class of adaptive optimal control problem that is more difficult than the pole-balancing problem. It has been investigated in Sutton (1996), where CMAC-based Sarsa-learning algorithms were employed to solve it and only the case of discrete control actions was studied. In our experiments, the case of continuous actions is considered.

An acrobot moving in the vertical plane is shown in Figure 15, where OA and AB are the first link and the second link, respectively. The control torque is applied at point A. The goal of the swing-up control is to swing the tip B of the acrobot above the line CD which is higher than the joint O by an amount of the length of one link.

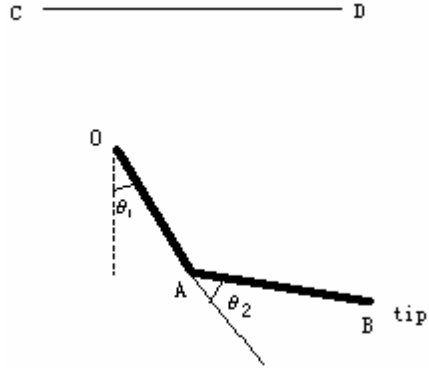

Figure 15: The acrobot

The dynamics of the acrobot system is described by the following equations.

$$\ddot{\theta}_1 = -(d_2\ddot{\theta}_2 + \phi_1)/d_1 \qquad (53)$$

$$\ddot{\theta}_2 = (\tau + d_2\phi_1/d_1 - \phi_2) \qquad (54)$$

where

$$d_1 = m_1 l_{c1}^2 + m_2(l_1^2 + l_{c2}^2 + 2l_1 l_{c2}\cos\theta_2) + I_1 + I_2 \qquad (55)$$

$$d_2 = m_2(l_{c2}^2 + l_1 l_{c2}\cos\theta_2) + I_2 \qquad (56)$$

$$\phi_1 = -m_2 l_1 l_{c2}\dot{\theta}_2^2\sin\theta_2 - 2m_2 l_1 l_{c2}\dot{\theta}_1\dot{\theta}_2\sin\theta_2 + (m_1 l_{c1} + m_2 l_1)g\cos(\theta_1 - \pi/2) + \phi_2 \qquad (57)$$

$$\phi_2 = m_2 l_{c2}g\cos(\theta_1 + \theta_2 - \pi/2) \qquad (58)$$

In the above equations, the parameters $\theta_i$, $\dot{\theta}_i$, $m_i, l_i, I_i, l_{ci}$ are the angle, the angle velocity, the mass, the length, the moment of inertia and the length of the center of mass for link $i$ ($i$=1,2), respectively.

Let $s_T$ denote the goal state of the swing-up control. Since the control aim is to swing up the acrobot in minimum time, the reward function $r_t$ is defined as

$$r_t = \begin{cases} 1, & \text{if } s = s_T \\ 0, & \text{else} \end{cases} \qquad (59)$$

In the simulation experiments, the control torque $\tau$ is continuous and is bounded by [-3$N$, 3$N$]. Similar to the cart-pole balancing problem, CMAC neural networks are applied to solve the above learning control problem with continuous states and actions. In the CMAC-based actor-critic controller, the actor network and the critic network both have $C$=4 tilings and $M$=7 partitions for each input. In the actor network, uniform coding is employed and non-uniform coding is used in the critic network. For details of the coding parameters, please refer to Appendix C. The sizes of the physical memories for the actor network and the critic network are 100 and 80, respectively. In the CMAC networks, the following hashing techniques are used. (For the definition of $A(s),a(i)$ and $F(s)$, please refer to Subsection 5.2.)

$$A(s) = \sum_{i=1}^{4}[a(i) \times M^{i-1}] \tag{60}$$

$$F(s)=A(s) \bmod K \tag{61}$$

In the simulation, the parameters for the acrobot are chosen as $m_1$=$m_2$=1kg, $I_1$=$I_2$=1kgm$^2$, $l_{c1}$=$l_{c2}$=0.5m, $l_1$=$l_2$=1m and g=9.8m/s$^2$. The time step for simulation is 0.05s and the time interval for learning control is 0.2s. The learning parameters are λ=0.6, γ=0.90, β=0.2, $k_1$=0.4, $k_2$=0.5. A trial is defined as the period that starts from the stable equilibrium and ends when the goal state is reached. After each trial, the state of the acrobot is re-initialized to its stable equilibrium. For each parameter setting, 5 independent runs are tested. Each run consists of 50 trials and after 50-th trial, the actor network is tested by controlling the acrobot alone, i.e., by setting the action variance defined in (43) to zero. The performance of the algorithms is evaluated according to the steps used by the actor networks to swing up the acrobot.

The performance comparisons between Fast-AHC and AHC are shown in Figure 16,17 and 18. In the experiments, both algorithms are tested with different λ and AHC is also tested with different learning factors of the critic networks.

From the results, it is also shown that Fast-AHC can achieve higher data efficiency than AHC. However, in this example, a relatively large $\delta$ is used, which is different from the previous cart-pole balancing example. In other experiments, good performance is obtained with large initializing constant and when $\delta$ is very small, the performance deteriorates significantly. Thus this problem may be referred to the low SNR case in Moustakides (1997), where large values of $\delta$ are preferable for best convergence rate of RLS methods.

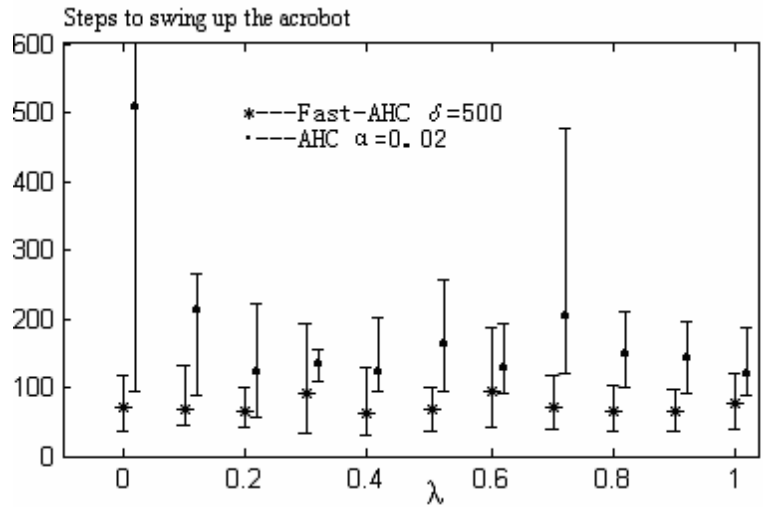

Figure 16: Performance comparison between Fast-AHC and AHC with α=0.02

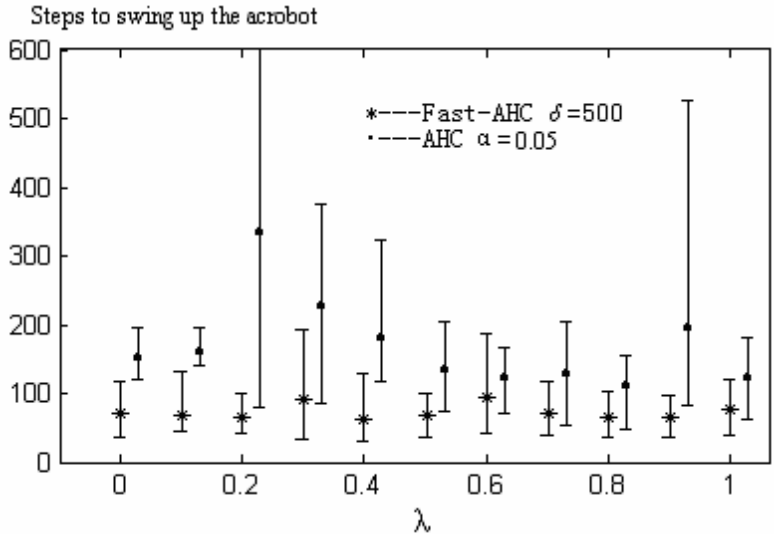

Figure 17: Performance comparison between Fast-AHC and AHC with α=0.05

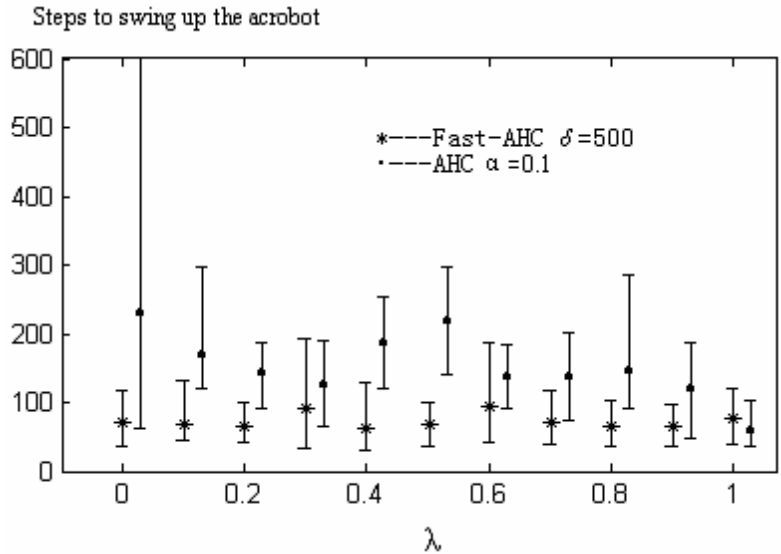

Figure 18: Performance comparison between Fast-AHC and AHC with α=0.1

The following Figure 19 shows the performance comparison between Fast-AHC with a large (300) and a small (0.01) value of $\delta$, where 6 settings of the parameter λ are tested for each algorithm. The performance of AHC using LS-TD(λ) is also shown. In Figure 20, a typical curve of the angle of the first link is plotted, where the acrobot is controlled by the actor network of the Fast-AHC method (λ=0.6) after 50 trials.

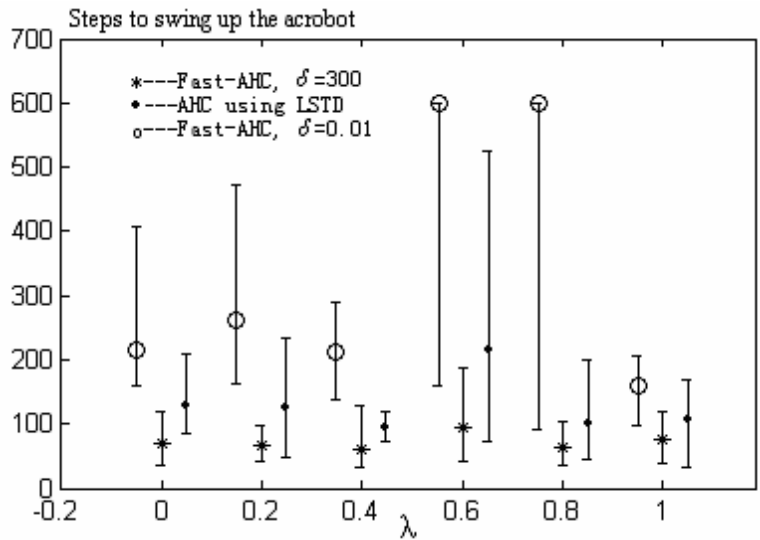

Figure 19: Performance comparison of Fast-AHC and AHC using LS-TD($\lambda$)

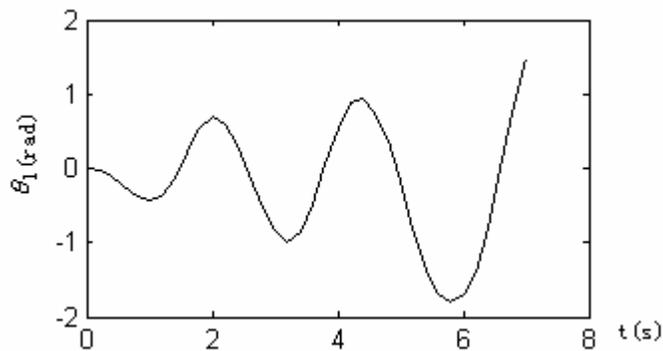

Figure 20: Variation of the angle of link 1(Controlled by Fast-AHC after 50 trials)

### 5.4 Analysis of The Experimental Results

Based on the above experimental results, it can be concluded that by using the RLS-TD($\lambda$) algorithm in the critic network, the Fast-AHC algorithm can obtain better performance than conventional AHC algorithms in that less trials or data are needed to converge to a near optimal policy. As is well known, one difficulty for the applications of RL methods is their slow convergence, especially in the cases where learning data are hard to be generated. For the Fast-AHC algorithm, although more computation per step is required than conventional AHC methods, it will not be a serious problem when the number of linear state features is small. In all of our learning control experiments, hashing techniques are used to reduce the state features in CMAC networks so that the computation of Fast-AHC can be reduced to an economical amount. Nevertheless, when the state feature number is large, conventional AHC methods may be preferable.

In the experiments, it is observed that the performance of Fast-AHC is affected by the initializing constant $\delta$. These results are consistent with the property of RLS-TD($\lambda$) and the RLS method in adaptive filtering, which has been discussed in Section 4. In the learning control experiments of the cart-pole balancing problem, better performance of Fast-AHC is obtained by using small values of $\delta$. While in the learning control of the acrobot, higher data efficiency is achieved using Fast-AHC with a relatively large $\delta$. These two different properties of Fast-AHC may be referred to the different SNR cases for RLS methods (Moustakides,1997). A thorough theoretical analysis on this problem is an interesting topic for future research.

In our experiments, the performance of the AHC method using LS-TD($\lambda$) is also tested. As has been studied in Section 4, when the initializing constant $\delta$ is large, the performance of RLS-TD($\lambda$) and LS-TD($\lambda$) does not differ much. So the performance of AHC using LS-TD($\lambda$) is similar to that of Fast-AHC with large values of $\delta$.

As studied in Moustakides (1997), the RLS method can converge much faster than other adaptive filtering methods if the environment is stationary and the initializing constant is selected appropriately. In some cases, RLS may converge almost instantly. This is also verified in the learning prediction experiments of the RLS-TD($\lambda$) algorithm. When applying RLS-TD($\lambda$) in an actor-critic learning controller, although the policy of the actor will change over time, it can still be assumed that the changing speed of the policy is slow when compared with the fast convergence speed of RLS-TD($\lambda$). Thus good performance of learning prediction can be obtained in the critic. Moreover, since the learning prediction performance of the critic is important to the policy learning of the actor, the improvement in learning prediction efficiency will contribute to the whole performance improvement of the controller.

## 6. Conclusions and Future Work

Two new reinforcement learning algorithms using RLS methods, which are called RLS-TD($\lambda$) and Fast-AHC, respectively, are proposed in this paper. RLS-TD($\lambda$) can be used to solve learning prediction problems more efficiently than conventional linear TD($\lambda$) algorithms. The convergence with probability 1 is proved for RLS-TD($\lambda$) and the limit of convergence is also analyzed. Experimental results on learning prediction problems show that the RLS-TD($\lambda$) algorithm is superior to conventional TD($\lambda$) algorithms in data efficiency and it also eliminates the design problem of the step sizes in linear TD($\lambda$) algorithms. RLS-TD($\lambda$) can be viewed as the extension of RLS-TD(0) from $\lambda=0$ to general $0<\lambda\leq1$. Although the effect of $\lambda$ on the convergence speed of RLS-TD($\lambda$) may not be significant in some cases, the usage of $\lambda>0$ will still affect the approximation error bound. Thus, when there are needs for value function estimation with high precision, large values of $\lambda$ are preferable to $\lambda=0$. Furthermore, RLS-TD($\lambda$) is superior to LS-TD($\lambda$) in computation when the weight vector must be updated after every observations.

Since learning prediction can be viewed as a sub-problem of learning control, we extend the results in learning prediction to a learning control method called the AHC algorithm. Using RLS-TD($\lambda$) in the critic network, Fast-AHC can achieve better performance than conventional AHC method in data efficiency. Simulation results on the learning control of the pole-balancing problem and the acrobot system confirm the above analyses.

In the experiments, it is found that the performance of RLS-TD($\lambda$) as well as Fast-AHC is influenced by the initializing constant $\delta$ of RLS methods. Different values of $\delta$ are needed for best performance in different cases. This is also a well-known phenomenon in RLS-based adaptive filtering and the theoretical results in Moustakides (1997) provide some basis for the explanations of our results. A complete investigation of this problem is our ongoing work.

The idea of using RLS-TD($\lambda$) in the critic network may be applied to other reinforcement learning methods with actor-critic architectures. In Konda and Tsitsiklis (1998), a new actor-critic algorithm using linear function approximators is proposed and the convergence under certain conditions is proved. One condition for the convergence of this algorithm is that the convergence rate of the critic is much faster than that of the actor. Thus the application of RLS-TD($\lambda$) in the critic may be preferable in order to ensure the convergence of the algorithm. The theoretical and empirical work on this problem deserves to be studied in the future.

### Acknowledgements

This work is supported by the National Natural Science Foundation of China under Grants 60075020, 60171003 and China University Key Teacher's Fellowship. We would very much like to thank the anonymous reviewers and Associate Editor Michael L. Littman for their insights and constructive criticisms, which have helped improve the paper significantly.

## Appendix A. Derivation of the RLS-TD($\lambda$) Algorithm

For the derivation of RLS-TD($\lambda$), there are two different cases, which are determined by the value of the forgetting factor.

(1) RLS-TD($\lambda$) with a unit forgetting factor.

Since

$$P_t = A_t^{-1} \tag{62}$$

$$P_0 = \delta I \tag{63}$$

$$K_{t+1} = P_{t+1}\vec{z}_t \tag{64}$$

According to Lemma 1,

$$
\begin{aligned}
P_{t+1} &= A_{t+1}^{-1} \\
&= P_t - P_t\vec{z}_t[1 + (\phi^T(x_t) - \gamma\phi^T(x_{t+1}))P_t\vec{z}_t]^{-1}(\phi^T(x_t) - \gamma\phi^T(x_{t+1}))P_t
\end{aligned}
\tag{65}
$$

$$
\begin{aligned}
K_{t+1} &= P_{t+1}\vec{z}_t \\
&= P_t\vec{z}_t / (1 + (\phi^T(x_t) - \gamma\phi^T(x_{t+1}))P_t\vec{z}_t)
\end{aligned}
\tag{66}
$$

$$
\begin{aligned}
W_{t+1} &= A_{t+1}^{-1}b_{t+1} \\
&= P_{t+1}(\sum_{i=0}^{t}\vec{z}_i r_i) \\
&= P_{t+1}(P_t^{-1}W_t + \vec{z}_t r_t)
\end{aligned}
\tag{67}
$$

Thus

$$
\begin{aligned}
W_{t+1} &= P_{t+1}[(P_{t+1}^{-1} - \vec{z}_t(\phi^T(x_t) - \gamma\phi^T(x_{t+1})))W_t + \vec{z}_t r_t] \\
&= W_t + P_{t+1}(\vec{z}_t r_t - \vec{z}_t(\phi^T(x_t) - \gamma\phi^T(x_{t+1}))W_t) \\
&= W_t + K_{t+1}[r_t - (\phi^T(x_t) - \gamma\phi^T(x_{t+1}))W_t]
\end{aligned}
\tag{68}
$$

(2) RLS-TD($\lambda$) with a forgetting factor $\mu < 1$

The derivation of RLS-TD($\lambda$) with a forgetting factor $\mu < 1$ is similar to the exponentially weighted RLS algorithm in Haykins (1996, pp.566-569). Here we only present the results:

$$K_{t+1} = P_t\vec{z}_t / (\mu + (\phi^T(x_t) - \gamma\phi^T(x_{t+1}))P_t\vec{z}_t) \tag{69}$$

$$W_{t+1} = W_t + K_{t+1}(r_t - (\phi^T(x_t) - \gamma\phi^T(x_{t+1}))W_t) \tag{70}$$

$$P_{t+1} = \frac{1}{\mu}[P_t - P_t\vec{z}_t[\mu + (\phi^T(x_t) - \gamma\phi^T(x_{t+1}))P_t\vec{z}_t]^{-1}(\phi^T(x_t) - \gamma\phi^T(x_{t+1}))P_t] \tag{71}$$

## Appendix B. Proof of Theorem 1

To study the steady property of the Markov chain defined in Section 3, we construct a stationary process as follows. Let $\{x_t\}$ be a Markov chain that evolves according to the transition matrix $P$ and is already in its steady state, which means that $\Pr\{x_t=i\}=\pi(i)$ for all $i$ and $t$. Given any sample path of the Markov chain, we define

$$\vec{z}_t = \sum_{\tau=-\infty}^{t} (\gamma\lambda)^{t-\tau} \phi(x_\tau) \tag{72}$$

Then $X_t = \{x_t, x_{t+1}, \vec{z}_t\}$ is a stationary process, which is the same as that discussed in (Tsitsiklis and Roy, 1997).

Let $D$ denote a $N \times N$ diagonal matrix with diagonal entries $\pi(1)$, $\pi(2)$,..., $\pi(N)$, where $N$ is the cardinality of state space $X$. Then Lemma 2 can be derived as follows.

**Lemma 2.** (Tsitsiklis and Roy, 1997) *Under Assumption 1-4, the following equations hold.*

$$1) \quad E_0[\phi(x_t)\phi(x_{t+m})] = \Phi^T DP^m \Phi \quad, \text{ for } m>0 \tag{73}$$

$$2) \quad E_0[\vec{z}_t\phi^T(x_t)] = \sum_{m=0}^{\infty} (\gamma\lambda)^m \Phi^T DP^m \Phi \,, \tag{74}$$

$$3) \quad E_0[\vec{z}_t r_t(x_t,x_{t+1})] = \sum_{m=0}^{\infty} (\gamma\lambda)^m \Phi^T DP^m \vec{r} \tag{75}$$

*where* $\vec{r} \in R^N$*, whose Nth component is equal to* $E[r(x_t,x_{t+1})|x_t=i]$*.*

According to Lemma 2, $E_0[A(X_t)]$ and $E_0[b(X_t)]$ are well defined and finite. Furthermore, $E_0[A(X_t)]$ is negative definite, so it is invertible.

From equation (67),

$$\begin{aligned} W_{RLS-TD(\lambda)} &= [P_0^{-1} + \sum_{t=1}^{T} A(X_t)]^{-1}[P_0^{-1}W_0 + \sum_{t=1}^{T} b(X_t)] \\ &= [\frac{1}{T}P_0^{-1} + \frac{1}{T}\sum_{t=1}^{T} A(X_t)]^{-1}[\frac{1}{T}P_0^{-1}W_0 + \frac{1}{T}\sum_{t=1}^{T} b(X_t)] \end{aligned} \tag{76}$$

Since

$$E_0[A(X_t)] = \lim_{T\to\infty} \frac{1}{T}\sum_{t=1}^{T} A(X_t) \tag{77}$$

$$E_0[b(X_t)] = \lim_{T\to\infty} \frac{1}{T}\sum_{t=1}^{T} b(X_t) \tag{78}$$

and $E_0[A(X_t)]$ is invertible,

$$\lim_{T\to\infty} W_{RLS-TD(\lambda)} = E_0^{-1}[A(X_t)]E_0[b(X_t)] = W^* \tag{79}$$

Thus $W_{RLS-TD(\lambda)}$ converges to $W^*$ with probability 1.

## Appendix C. Some details of the coding structures of CMAC networks

In the following discussion, the coding structures of CMAC networks in the cart-pole balancing problem and the acrobot control problem are presented.

**(1) CMAC coding structures in the cart-pole balancing problem**

In the CMAC networks, the state variables have the following boundaries.

$$\theta \in [-12°, 12°], \qquad \dot{\theta} \in [-50 \deg/s, 50 \deg/s]$$

$$x \in [-2.4, 2.4], \qquad \dot{x} \in [-1,1]$$

For the critic network, $C$=4 and $M$=7. The hashing technique specified in equations (50) and (51) is employed and the total memory size is 30.

For the actor network, $C$=4 and $M$=7. The hashing technique specified in equations (60) and (61) is employed and the total memory size is 100.

**(2) CMAC coding structures in the acrobot swing-up problem**

In the simulation, the angles are bounded by $[-\pi, \pi]$ and the angular velocities are bounded by $\dot{\theta}_1 \in [-4\pi, 4\pi]$, $\dot{\theta}_2 \in [-9\pi, 9\pi]$. The tiling numbers of the actor and the critic both are equal to 4 ($C$=4). The total memory sizes for the critic and the actor are 80 and 100, respectively. In the actor network, each tiling partitions the range of each input into 7 equal intervals ($M$=7). In the critic network, the partitions for each input are non-uniform, which are given by

$\theta_1 : \{ -\pi, -1, -0.5, 0, 0.5, 1, \pi \}, \qquad \dot{\theta}_1 : \{-4\pi, -1.5\pi, -0.5\pi, 0, 0.5\pi, 1.5\pi, 4\pi\}$

$\theta_2 : \{-\pi, -1, -0.5, 0, 0.5, 1, \pi \}, \qquad \dot{\theta}_2 : \{-9\pi, -2\pi, -0.5\pi, 0, 0.5\pi, 2\pi, 9\pi\}$

## References

Albus,J.S.(1975). A new approach to manipulator control: the cerebellar model articulation controller (CMAC). *Journal of Dynamic Systems, Measurement, and Control*, 97(3), 220-227.

Barto,A.G., Sutton R.S., & Anderson C.W. (1983). Neuronlike adaptive elements that can solve difficult learning control problems. *IEEE Transactions on System, Man, and Cybernetics,*13, 834-846.

Bertsekas D.P. & Tsitsiklis J.N. (1996). *Neurodynamic Programming*. Belmont, Mass.: Athena Scientific.

Berenji H.R. & Khedkar P. (1992). Learning and tuning fuzzy logic controllers through re-inforcements, *IEEE Trans.On Neural Networks*, 3(5), 724-740.

Boyan. J.(1999). Least-squares temporal difference learning. In Bratko, I., and Dzeroski, S., eds., *Machine Learning*: *Proceedings of the Sixteenth International Conference* (ICML).

Boyan, J.(2002). Technical update: least-squares temporal difference learning. *Machine Learning, Special Issue on Reinforcement Learning*, to appear.

Brartke. S.J. & Barto A. (1996). Linear least-squares algorithms for temporal difference learning. *Machine Learning,* 22, 33-57.

Dayan P.(1992). The convergence of TD($\lambda$) for general $\lambda$. *Machine Learning*, 8, 341-362.

Dayan P.. & Sejnowski T.J. (1994). TD($\lambda$) converges with probability 1. *Machine Learning,* 14, 295-301.

Eleftheriou E. & Falconer,D.D. (1986). Tracking properties and steady state performance of RLS adaptive filter algorithms. *IEEE Transactions on Acoustics, Speech, and Signal Processing*, 34, 1097-1110.

Eweda E. & Macchi, O. (1987). Convergence of the RLS and LMS adaptive filters. *IEEE Trans. Circuits and Systems*, 34, 799-803.

Haykin S. (1996), *Adaptive Filter Theory*, 3[rd] edition, Englewood Cliffs, NJ: Prentice-Hall.

Hubing N.E. & Alexander S.T. (1989). Statistical analysis of the soft constrained initialization of RLS algorithms. In *Proc. of the IEEE International Conference on Acoustics, Speech and Signal Processing*.

Jaakkola T., Jordan M.I., & Singh S.P. (1994). On the convergence of stochastic iterative dynamic programming algorithms. *Neural Computation*. 6(6), 1185-1201.

Kaelbling L.P., Littman M.L., & Moore A.W. (1996). Reinforcement learning: a survey. *Journal of Artificial Intelligence Research*, 4, 237-285.

Konda V.R, & Tsitsiklis J.N. (2000). Actor-critic algorithms. In *Neural Information Processing Systems*, 2000, MIT Press.

Lin L.J. (1992). Self-improving reactive agents based reinforcement learning, planning and teaching. *Machine Learning*, 8(3/4), 293-321.

Lin C.T. & Lee C.S.G. (1994). Reinforcement structure/parameter learning for neural-network-based fuzzy Logic control system. *IEEE Transactions on Fuzzy System*, 2(1), 46-63.

Ljung L. & Soderstron T. (1983). *Theory and Practice of Recursive Identification*. MIT Press.

Ljung L. (1977). Analysis of recursive stochastic algorithm. *IEEE. Transactions on Automatic Control*, 22, 551.

Michie D. & Chambers R.A. (1968). BOXES: An experiment in adaptive control. *Machine Intelligence* 2, Dale E. and Michie D., eds., Edinburgh: Oliver and Boyd, 137-152.

Minsky M.L. (1954). Theory of neural-analog reinforcement systems and its application to the brain-model problem. Ph.D. Thesis, Princeton University.

Moustakides G.V. (1997). Study of the transient phase of the forgetting factor RLS. *IEEE Trans. on Signal Processing*, 45(10), 2468-2476.

Samuel A.L. (1959). Some studies in machine learning using game of checkers. *IBM Journal on Research and Development*, 3, 211-229.

Singh, S.P., Jaakkola T., Littman M.L., & Szepesvari C. (2000). Convergence results for single-step on-policy reinforcement-learning algorithms. *Machine Learning*, 38, 287-308.

Sutton R. & Barto A. (1998). *Reinforcement Learning, an Introduction.* Cambridge MA, MIT Press.

Sutton R. (1988). Learning to predict by the method of temporal differences. *Machine Learning*, 3(1), 9-44.

Tsitsiklis J.N. (1994). Asynchronous stochastic approximation and Q-learning. *Machine Learning*, 16, 185-202.

Tsitsiklis J.N. & Roy B.V. (1994). Feature-based methods for large scale dynamic programming. *Neural Computation*. 6(6), 1185-1201.

Tsitsiklis J.N. & Roy B.V. (1997). An analysis of temporal difference learning with function approximation. *IEEE Transactions on Automatic Control*. 42(5), 674-690.

Watkins C.J.C.H. & Dayan P. (1992). Q-Learning. *Machine Learning*. 8, 279-292.

Young P. (1984). *Recursive Estimation and Time-Series Analysis*. Springer-Verlag.